\pdfoutput=1

\documentclass[11pt]{article}

\usepackage{acl}

\usepackage{times}
\usepackage{latexsym}

\usepackage[T1]{fontenc}

\usepackage[utf8]{inputenc}

\usepackage{microtype}

\usepackage{inconsolata}

\usepackage{adjustbox}
\usepackage{graphicx}
\usepackage{enumitem}
\usepackage{url}
\usepackage{booktabs}
\usepackage{makecell}
\usepackage{amsmath}
\usepackage{multirow}
\usepackage{pgfplots}
\usepackage{pgfplotstable}
\usepackage{listings}
\usepackage{csquotes}

\title{\ours: Iterative Planning in Textual Environments}

\author{Li Zhang$^{1}$\thanks{\hspace{1mm} Work done as an intern at AI2.} \quad
  Peter Jansen$^{3}$ \quad Tianyi Zhang$^{1}$
   \\
  \textbf{Peter Clark}$^{2}$ \quad \textbf{Chris Callison-Burch}$^1$ \quad
  \textbf{Niket Tandon}$^2$ \\
  $^1$University of Pennsylvania\quad\quad $^2$Allen Institute for Artificial Intelligence \\
  $^3$The University of Arizona \\
  {\tt \{zharry\}@upenn.edu} \quad \tt{\{nikett\}@allenai.org}
}

\usepackage{tikz}
\usepackage{pgfplots}
\usepackage{xspace}

\newcommand{\ours}{\textsc{pddlego}\xspace}

\usepackage{soul}

\newcommand{\squishlist}{
  \begin{list}{$\bullet$}
    { \setlength{\itemsep}{0pt}      \setlength{\parsep}{3pt}
      \setlength{\topsep}{3pt}       \setlength{\partopsep}{0pt}
      \setlength{\leftmargin}{1.5em} \setlength{\labelwidth}{1em}
      \setlength{\labelsep}{0.5em} } }
\newcommand{\reallysquishlist}{
  \begin{list}{$\bullet$}
    { \setlength{\itemsep}{0pt}    \setlength{\parsep}{0pt}
      \setlength{\topsep}{0pt}     \setlength{\partopsep}{0pt}
      \setlength{\leftmargin}{0.2em} \setlength{\labelwidth}{0.2em}
      \setlength{\labelsep}{0.2em} } }

 \newcommand{\squishend}{
     \end{list} 
 }

\renewcommand{\cite}{\citep}

\definecolor{lightgray}{gray}{0.9}
\definecolor{Box1Color}{RGB}{227, 236, 246}
\definecolor{Box2Color}{RGB}{248, 220, 225}
\definecolor{Box3Color}{RGB}{255, 238, 224}
\definecolor{cbBlue}{RGB}{0, 114, 178}
\definecolor{cbOrange}{RGB}{240, 228, 66}
\definecolor{cbGreen}{RGB}{0, 158, 115}
\definecolor{cbRed}{RGB}{213, 94, 0}
\definecolor{cbPurple}{RGB}{204, 121, 167}
\definecolor{cbSkyBlue}{RGB}{86, 180, 233}
\definecolor{cbGray}{RGB}{128, 128, 128}
\definecolor{CBF1}{RGB}{255,99,132}  
\definecolor{CBF2}{RGB}{54,162,235}  
\definecolor{CBF3}{RGB}{255,206,86}  
\definecolor{CBF4}{RGB}{75,192,192}  
\definecolor{CBF5}{RGB}{153,102,255} 
\definecolor{CBF1b}{RGB}{205,89,112}  
\definecolor{CBF2b}{RGB}{44,142,215}  
\definecolor{CBF5b}{RGB}{133,92,225}  

\begin{document}
\maketitle
\begin{abstract}
Planning in textual environments have been shown to be a long-standing challenge even for current models. A recent, promising line of work uses LLMs to generate a formal representation of the environment that can be solved by a symbolic planner. However, existing methods rely on a fully-observed environment where all entity states are initially known, so a one-off representation can be constructed, leading to a complete plan. In contrast, we tackle partially-observed environments where there is initially no sufficient information to plan for the end-goal. We propose \ours that \textbf{iteratively} construct a planning representation that can lead to a partial plan for a given sub-goal. By accomplishing the sub-goal, more information is acquired to augment the representation, eventually achieving the end-goal. We show that plans produced by few-shot \ours are 43\% more efficient than generating plans end-to-end on the Coin Collector simulation, with strong performance (98\%) on the more complex Cooking World simulation where end-to-end LLMs fail to generate coherent plans (4\%).\footnote{Our code can be found at \url{https://github.com/zharry29/nl-to-pddl}.}
\end{abstract}

\section{Introduction}
Planning with LLMs has witnessed a surge of interest in the NLP community, not only because it showcases AI systems' ability to reason about complex events, but also because of the need of many downstream applications like goal-driven robotics \cite{huang2022language,huang2022inner} and intelligent planning assistants \cite{lyu-etal-2021-goal}. The most intuitive approach of this task is using LLMs as planners to produce a sequence of actions executed to arrive at a goal state \cite{valmeekam2023planbench,stein2023autoplanbench}. While applicable in many domains, this LLM-based approach is found to underperform in textual simulated environments \cite{valmeekam2023large,valmeekam2023planning} and to lack interpretability compared to symbolic planning methods that derive a plan from a formal representation of the environment. We join the efforts that combine both approaches, effectively translating the textual input into a symbolic representation expressed in the planning domain definition language (PDDL) (see Appendix~\ref{app:pddl} for an introduction), which can then be solved by a symbolic planner \cite{collins2022structured,lyu2023faithful,liu2023llm+,xie2023translating,wong2023learning}. This neurosymbolic approach has gained popularity as it combines LLMs' flexibility to understand rich NL and classical planners' determinism and faithfulness. 

\begin{figure}
    \centering
    \includegraphics[width=0.9\columnwidth]{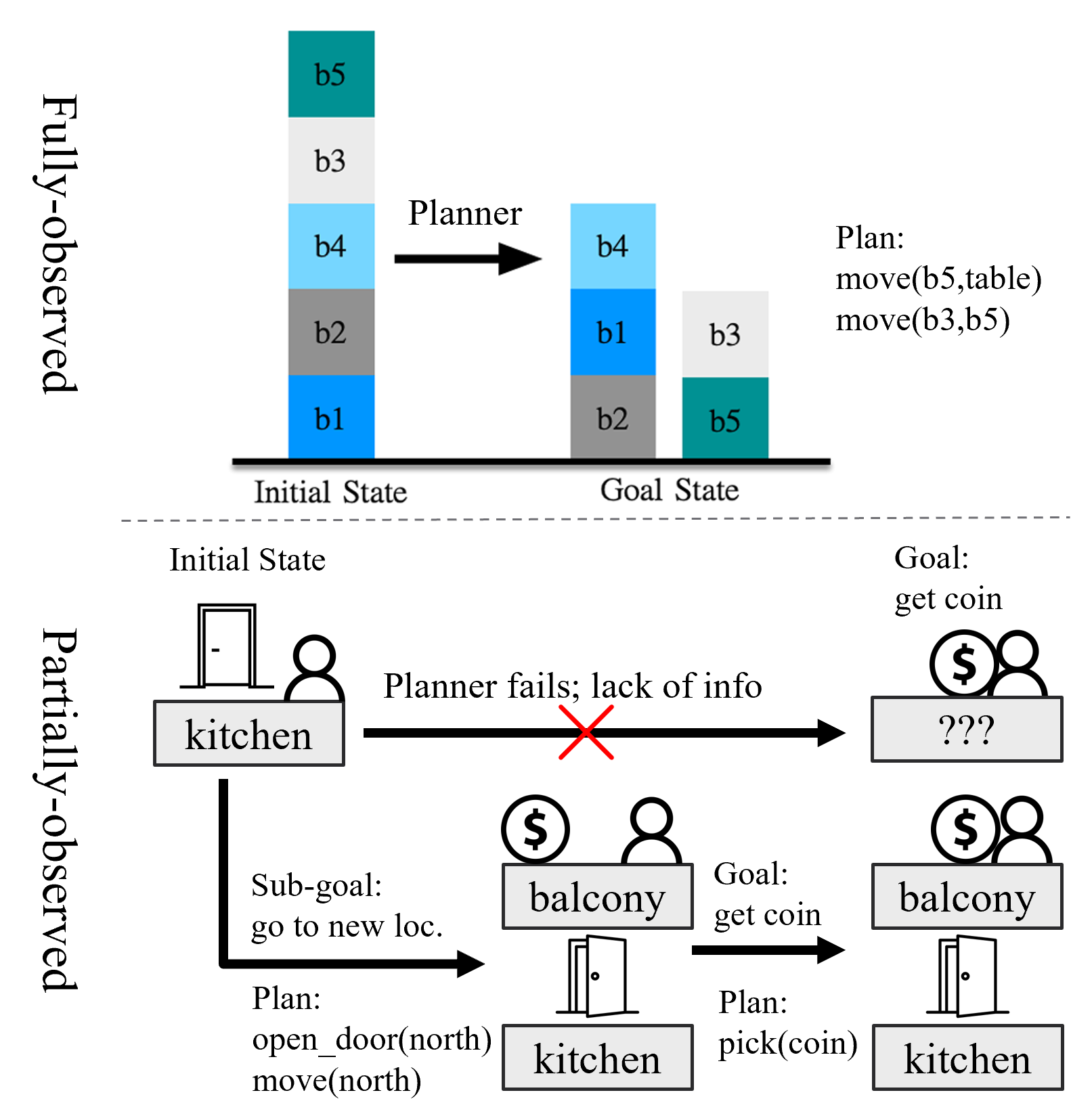}
    \caption{A fully-observed environment like BlocksWorld (upper, to rearrange objects from and to a given configuration) can be tackled by generating a PDDL problem file, while a partially observed one like Coin Collector (lower, to look for an object in an unknown location) cannot until sufficient exploration.}
    \label{fig:fully_partially_observed}
\end{figure}

All previous work on LLM generating PDDL has only experimented on \textbf{fully-observed} environments where all entity states are initially known, thus requiring no exploration. Take BlocksWorld, a common benchmark for such work, as an example (Figure~\ref{fig:fully_partially_observed}, upper), both the initial and goal states are initially spelled out, in which case the LLM's job is akin to translating the textual descriptions of the environment into a PDDL problem file which specifies the initial and goal entity states. Assuming also a domain file, a one-off plan can be found and executed to reach the end-goal. In contrast, many real-world environments are \textbf{partially-observed} (Figure~\ref{fig:fully_partially_observed}, lower), where the entity states dynamically get uncovered during exploration. Since the necessary initial and goal states might also be unknown (e.g., looking for an item without knowing where it is), the previous approach falls apart due to the impossibility to specify a complete problem file. This causes a chicken-and-egg problem where a plan is required for exploration, while exploration is required to build PDDL that results in a plan. Given this challenge, past work on partially-observed environments has only used LLMs to directly generate plans \cite{shinn2023reflexion,majumder2023clin}, but not a planning representation.

To break the above stalemate, we propose \ours, a methodology to use LLMs to iteratively build a PDDL problem file from textual observations from the environment. In this problem file, the initial states (or rather current states) reflect the current knowledge of the environment, while the goal states can be dynamically adjusted. In case the problem file does not contain sufficient information to plan for the end-goal (e.g., find a coin), \ours recursively falls back to a provided sub-goal (e.g., go to an unvisited room). This way, a plan can be found to reach the sub-goal, leading to new observations by exploring the environment, and iteratively refine the problem file until a plan can be found for the end-goal. 

We evaluate \ours on benchmarks of textual interactive virtual environments akin to the robotic planning simulations where PDDL is known for. Our PDDL-induced plans are 43\% more efficient than LLMs generating plans directly on the Coin Collector simulation. On one setting of the more complex Cooking World simulation \ours achieves near-perfect 98\% success rate where LLMs that predict action achieves only 4\%, while on a more challenging setting, 46\% over 0\%. 

\begin{figure}
    \centering
    \includegraphics[width=0.9\columnwidth]{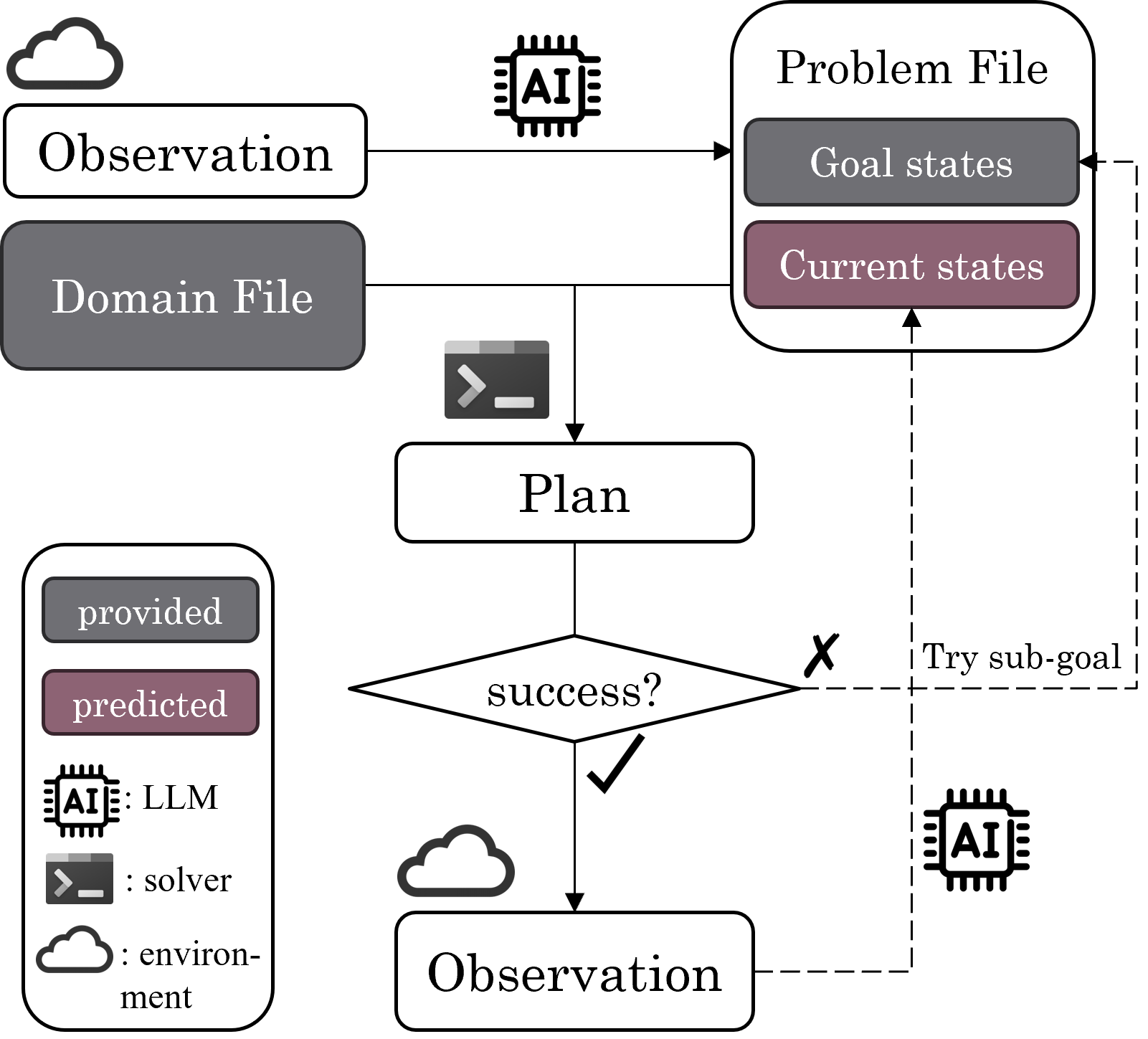}
    \caption{The pipeline of \ours. A PDDL problem file is iteratively built during exploration.}
    \label{fig:pipeline}
\end{figure}

\section{Methodology}
Our approach is illustrated in Figure~\ref{fig:pipeline}. We operate in a partially-observed, textual, simulated environment which functions as a multi-turn interaction between the environment and the agent (e.g., \textit{a game to find an item}). Specifically, the environment provides an observation (\textit{objects in a room}) along with a list of permitted actions (\textit{move, pick up}). Then, the agent selects on of these actions, and repeats. The environment can be seen as a finite state machine where each state consists of the conjunction of all entity states and determines the permitted actions. The agent succeeds when a goal state is reached (\textit{the sought item is in hand}); it fails when it cannot possibly reach goal state.

Like most prior work in using LLMs to generate a planning representation like PDDL, we assume that a domain file that defines the available actions is provided; this domain file can solve a problem file that defines the initial and goal entity states (\textit{where the agent is, where the item is, how are these two locations connected}) when possible to result in a plan (\textit{go west, pick up item}). We also assume a sub-goal structure, namely, an array of goal states defined in PDDL that a model can fall back to when the current goal is unattainable. 

\setlength{\abovedisplayskip}{3pt}
\setlength{\belowdisplayskip}{3pt}

Formally, we are initially presented with the first observation $o_1$ with the end-goal $G$. We use an LLM to construct an initial problem file $PF_1$ (\{current states, goal states\}) to plan for this end-goal. 
\begin{equation}
    PF_1 = \{LLM(o_1), G\}
\end{equation}
If this problem file can be solved by the provided domain file with a solver, a plan containing one or more actions is found. 
\begin{equation}
    Plan_1 := (a_1^1, a_1^2, \dots) = solver(DF,PF_1)
\end{equation}
If a plan cannot be found due to a lack of information in the problem file, the goal $G$ is swapped out by an immediate sub-goal $G'$, and the solver retries. The actions in the plan are then sequentially executed in the current environment $E$, resulting in a list of new observations.
\begin{equation}
    (E, o_2^1, o_2^2, \dots) = exec(E, a_1^1, a_1^2, \dots)
\end{equation}
Thus begins the second iteration. Using the new observations, the previous problem file is regenerated (referred to as \textbf{PDDL-gen}).
\begin{equation}
    PF_2 = \{LLM(PF_1, o_2), G\}
\end{equation}
The process goes on until one observation fulfills the termination condition.

Unlike prior work that generates the problem file once, \ours's having LLMs iteratively generating the problem file often result in inconsistencies and errors (e.g., missing a connectivity relation between two rooms, using the name a room in a relation without declaring the room, missing a parenthesis, etc.). To tackle this, we have the LLMs only predict the change in the problem file (i.e., the change of entity states), which we deterministically applied to the previous problem file (referred to as \textbf{PDDL-edit}).
\begin{equation}\tag{4'}
    \Delta_{2} = LLM(PF_1, o_2),\ PF_2 = PF_1 + \Delta_{2}
\end{equation}

We will compare our two approaches above with the baseline where LLMs directly generate an action (referred to as \textbf{Action-gen}).
\begin{equation}\tag{2'}
    Plan_i = LLM(o_i)
\end{equation}

\section{Environments}
We experiment with two goal-oriented, partially-observed simulated environments, or text games, that span a variety of difficulty and flavor. 

\noindent \textbf{Coin Collector} \cite{yuan2019counting} focuses on navigation, which is an indispensable element of most simulations. The agent's task is to explore rooms, some connected by locked doors, and find a coin, similar to the running example above. Just as previously discussed, the previous approach on generating a PDDL problem file cannot be applied to Coin Collector because the location of the coin is unknown until the agent enters the same room as the coin. Therefore, the sub-goal structure for this tasks is defined as:
\begin{enumerate}[topsep=0pt,itemsep=-1ex,partopsep=1ex,parsep=1ex]
    \item pick up coin (requires the location of the coin)
    \item go to a room that has not been visited (reveals location of the coin)
\end{enumerate}

The sub-goal of ``going to an unvisited room'' results in monotonously increasing progress to the end-goal of ``finding the coin''. In similar search-related tasks, this singular sub-goal or strategy suffices, though it may not work for all situations.


\noindent \textbf{Cooking World} \cite{ijcai2020p207} subsumes Coin Collector with more complex tasks. The agent' task is to first explore rooms to find ingredients required by a recipe, much like Coin Collector. Next, it should cook the ingredient in some specified location using some specified appliance. Finally, when all ingredients are cooked correctly, a meal can be successfully prepared. Therefore, the sub-goal structure for this tasks is defined as:
\begin{enumerate}[topsep=0pt,itemsep=-1ex,partopsep=1ex,parsep=1ex]
    \item prepare meal (requires having obtained each ingredient and located relevant appliances)
    \item pick up each ingredient (requires the location of each ingredient; obtains ingredients)
    \item go to a room that has not been visited (reveals location of ingredients and appliances)
\end{enumerate}


\begin{table*}[]
\centering
\small
\begin{tabular}{l|l|lll|lll}
\toprule
         & \multicolumn{1}{l|}{random} & \multicolumn{3}{c|}{GPT 3.5 Turbo} & \multicolumn{3}{c}{GPT 4 Turbo} \\ \midrule
         & & Action-gen  & PDDL-gen$^\dag$  & PDDL-edit$^\dag$ & Action-gen    & PDDL-gen$^\dag$   & PDDL-edit$^\dag$   \\
Coin    & 4\%  & 68\% & 26\% & 28\% & \textbf{94\%} & 58\% & 78\%    \\
Cooking (easy) & 0\% & 0\% & 70\% & 68\% & 4\% & 94\% & \textbf{98\%} \\
Cooking (hard) & 0\% & 0\% & 4\%  & 6\% & 0\% & 16\% & \textbf{46\%}            \\\bottomrule
\end{tabular}
\caption{The percentage where the agent succeeds by taking no more than the maximum steps on the test set. The $^\dag$ sign specifies methods under our proposed \ours methodology. }
\label{tab:results}
\end{table*}

To better understand these simulations, example trajectories are shown in Appendix~\ref{app:trajectories}.

\section{Evaluation}
For both simulations, we use the implementation from \citet{jansen2022textworldexpress}. For Coin Collector, we use the most complex setting; for Cooking World, we consider an easy and a hard setting with varying number of locations and ingredients. See more details in Appendix~\ref{app:hyperparameters}. For the choice of LLM, we consider \texttt{gpt-3.5-turbo-1106} (GPT 3.5 Turbo) and \texttt{gpt-4-1106-preview} (GPT 4 Turbo) across baseline methods (i.e., Action-gen, PDDL-gen, and PDDL-edit). For Action-gen, we prompt the LLM with a full description of the simulation, and for PDDL methods, with a hand-annotated domain file containing well-defined actions. For the PDDL-edit setting, we prompt the LLM to generate templated edits (add, replace, and delete lines in the problem file). The prompt of each method include a 1-shot demonstration of the output format. See details of prompt design and domain file annotation in the Appendix~\ref{app:df}.

Regarding \textbf{performance}, Table~\ref{tab:results} shows a drastic performance degradation of Action-gen moving from Coin Collector (only 2 valid actions: move, open door each with 4 direction arguments) to the much more complex Cooking World (with 8 more actions with infinite possible arguments, like processing an ingredient). Moreover, in Cooking World, an agent would fail if an ingredient is processed incorrectly (e.g., fried instead of grilled, was not chopped before roasted). Therefore, LLMs generating actions on the fly are more likely to make irrevocable mistakes and fail the task. In contrast, our two-stage PDDL generation approaches ensure the correctness of the plan to process the ingredients (in the second stage) \textit{assuming} that the ingredients are gathered and that the appliances are identified (in the first stage). Logically, the failures of \ours indicates an inconsistency between the environmental observation and the problem file. For example, the connectivity of the rooms may not be updated correctly upon entrance to a new room, causing no plan or invalid plans to be found. By lessen the burden on LLMs, PDDL-edit notably ameliorates but cannot eliminate this issue. On Coin Collector, issues frequently arise in a loop, where opening a new door leads to a visited room. Notably, GPT3.5 is far worse than GPT4 in generating PDDL, in line with the observations by \citet{zhang-etal-2024-proc2pddl} and \citet{silver2023generalized}. 

Regarding \textbf{efficiency}, Figure~\ref{fig:coin_step_count} shows that on Coin Collector, PDDL-edit is no less efficient than Action-gen on 7 out 8 examples (red crosses are often lower than the blue circles) in the development set where PDDL-edit terminates successfully. Scaling up to the entire test set, with GPT4, PDDL-edit has an average step to success of 7.8 compared to Action-gen's 13.6 among successful attempts, a 43\% improvement on efficiency. Among these steps, 3.3 of Action-gen are invalid (e.g., moving through a closed door) compared to merely 0.2 of \ours, a significant difference when trials and errors are expensive. \ours also shows better \textbf{stability}. In Figure~\ref{fig:coin_step_count}, PDDL-edit exhibits a much smaller variance across runs than Action-gen. For example, if the coin happens to be immediately to the west of the initial room, deciding to go west initially would result in a prompt success, while exploring the east portion initially would result in a notable detour. Our approach of PDDL generation leaves only the task of parsing environmental configuration to the LLM, while the planning task is done deterministically by the solver, leading to more consistent plans across runs.

Regarding \textbf{interpretability} and \textbf{correctability}, the black-box nature of LLMs results in no faithful interpretation behind the decisions (c.f., thought-process). In Coin Collector, for example, if the coin has not be found at the maximum permitted steps, a problematic Action-gen trajectory is almost impossible to manually correct unless a human is to plot a map and keep track of the exploration. On the other hand, both PDDL-gen and PDDL-edit guarantees the correctness of the plan assuming that the generated or edited problem file is correct. Hence, upon failure, a human only needs to inspect and correct the most recent observation and the PDDL. For PDDL-edit, the job is even easier as only the change in the problem needs to be considered. An example learned problem file can be found in Appendix~\ref{app:example_pf}.

\pgfplotstableread{
x         y    y-max  y-min
0  35.4 6.6   6.6
1 25 12.2   12.2
2     17.9 16.6   16.6
3  0 0   0
4 9.6 6.8   6.8 
5     3 0   0
6  10.6 10.9   10.9
7 12 10.6   10.6
8     20.4 7.3   7.3
9  0 0   0
}{\plangen}

\pgfplotstableread{
x         y    y-max  y-min
1 7 0   0
2     3.4 0.9   0.9
3  0 0   0
4 7 0   0
5     3 0   0
7 1 0   0
8 25 9.2 9.2
9  0 0   0
}{\pddlgen}

\begin{figure}
    \centering
    \begin{tikzpicture}[scale=0.65] 
    \begin{axis} [symbolic x coords={0,1,2,3,4,5,6,7,8, 9},xtick=data, xlabel={example ID}, ylabel={num. steps to success}]
    \addplot[only marks, mark=o, blue] 
      plot[error bars/.cd, y dir=both, y explicit]
      table[x=x,y=y,y error plus expr=\thisrow{y-max},y error minus expr=\thisrow{y-min}] {\plangen};
    \addplot[only marks, mark=x, red] 
      plot[error bars/.cd, y dir=both, y explicit]
      table[x=x,y=y,y error plus expr=\thisrow{y-max},y error minus expr=\thisrow{y-min}] {\pddlgen};
    \legend{Action-gen,PDDL-edit}
    \end{axis}
    \end{tikzpicture}
    \caption{On Coin Collector, the mean and standard deviation of number of steps to success (less is better) for each development example, each over 5 trials with different random seeds of \texttt{gpt-4-1106-preview}, comparing Action-gen and PDDL-edit. The error bar represents the sample standard deviation. On example 0 and 6, PDDL-edit fails and thus not shown.}
    \label{fig:coin_step_count}
\end{figure}

\section{Conclusion}
We propose \ours, the first approach to use LLMs to iteratively learn a planning representation while exploring partially-observed environments. We quantitatively show the improvement of performance, efficiency and stability, while qualitatively argue the benefit of interpretability and correctability. Future work might remove the assumption of a domain file and a sub-goal structure.

\section*{Limitations}

Despite the many benefits of \ours, it also poses the following shortcomings compared to having LLMs directly generating the plan or actions. 

The first is speed and cost, as both the input and output become much longer to include PDDL code. For the OpenAI model we experiment with, PDDL-gen and PDDL-edit are on average about 5x slower than Action-gen. On the other hand, it is difficult to compare the cost which is highly dependent on prompt design. In our work, Action-gen keeps appending the chosen action, new observation and valid actions to the prompt, resulting in a longer input and higher cost for every exploration step. However, our PDDL methods only retain the most recent observation and problem file, so the input length, though initially longer, is roughly constant. 

The second is flexibility, which is the strong-suit of methods leveraging LLMs to do most of the work. For each environment we experiment with, a certain extent of hard-coding is required for our methods to work, hindering generalization. In our case, the domain file and sub-goals of one or more problem file for each environment must be manually annotated. Doing so presumes some prior insight into the environment, and therefore \ours is not truly a zero-shot methodology.

While the aim of this work is to show the preliminary gains of generating PDDL while exploring partially-observed environments, there could be stronger Action-gen baselines, such as using chain-of-thought to formulate a plan first instead of selecting actions on the fly, or more advanced methods in the literature. 

\section*{Acknowledgements}
This work is supported in part by the DARPA KAIROS Program (contract FA8750-19-2-1004), AFRL (contract FA8750-23-C-0507), the Office of the Director of National Intelligence (ODNI) via the IARPA HIATUS Program (contract 2022-22072200005), the NSF (Award 1928631), and gifts from Roblox and Salesforce. Approved for Public Release, Distribution Unlimited. The views and conclusions contained herein are those of the authors and should not be interpreted as necessarily representing the official policies, either expressed or implied, of DARPA, ODNI, IARPA, NSF, AFRL, the U.S. Government, or of Roblox or Salesforce. The U.S. Government is authorized to reproduce and distribute reprints for governmental purposes notwithstanding any copyright annotation therein.

\bibliography{anthology,custom}

\appendix


\section{Formulation of PDDL}
\label{app:pddl}

\begin{figure}[t]
    \centering
    \includegraphics[width=0.48\textwidth]{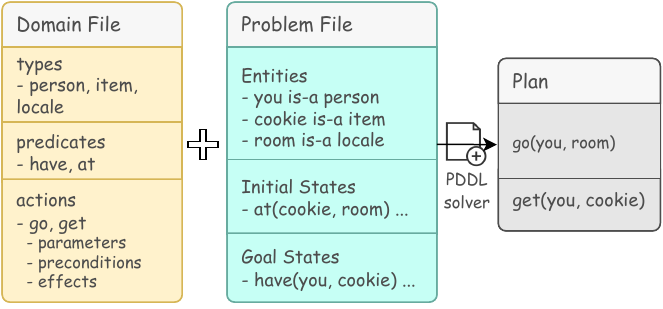}
    \caption{A PDDL solver produces a plan based on a minimal domain file and problem file. Previous work assumes the domain file as given, while we predict the action definitions in the domain file.}
    \label{fig:pddl_example}
\end{figure}

As shown in Figure~\ref{fig:pddl_example}, an instance of PDDL \cite{aeronautiques1998pddl} consists of a domain file, describing the actions, and a problem file, describing the initial and goal states of entities. A well-formed pair of domain and problem files can be solved by a symbolic planner, whose output is a sequence of actions.

\section{Annotated Domain Files and Prompts}
\label{app:df}
\ours is a method to iteratively construct problem files based on a provided domain file. Figure~\ref{fig:coin_df} and~\ref{fig:cooking_df} show the annotated domain files for Coin Collector and Cooking World, respectively. Note that the actions and their parameter lists in the domain file strictly maps to the permitted actions in the simulations, so that a PDDL plan can be mapped onto executable actions in the environment. Based on the domain file, our prompts for either generating (PDDL-gen) or editing (PDDL-edit) the problem file are simply (for Coin Collector):
\begin{displayquote}
\small
    You will continue to build a PDDL representation of an environment while exploring it. We will be using the following domain file: 
    <<domain file>>
    For example, for the given observation:

You are in the kitchen. To the South you see a closed wooden door.

Your task is to go to a location you have not been yet. You will generate the following problem file:
    <<example domain file>>
    
    Now, let's start afresh. 
\end{displayquote}
For PDDL-edit, a few more details are appended.
\begin{displayquote}
\small
    <<the above prompt>>
    
    Let's work with an example. Say you're given this observation:
You are in the kitchen. To the South you see a closed wooden door. To the East you see a closed glass door.

You will modify the above problem file using add, replace, and delete operations (in a JSON format). You SHOULD NOT provide a problem file directly.
\begin{lstlisting}
{
  "objects": {
    "add": [
      "loc1 - location",
      "loc2 - location"
    ],
    "replace": {},
    "delete": []
  },
  "init": {
    "add": [
      "(connected kitchen loc1 south)",
      "(closed_door kitchen loc1)",
      "(connected kitchen loc2 east)",
      "(closed_door kitchen loc2)"
    ],
    "replace": {},
    "delete": []
  }
}
\end{lstlisting}

Note a couple of things:

1. When you see a closed door, you would use a placeholder for the room behind the door.

2. When you enter a room, you learn the name of the room and will replace the placeholder with the name. You should also make sure to replace that name for all relations under "init".

3. When you enter a room, you're "at" the room and it becomes "visited". You should also delete other "at" conditions because you can only be at one room.

4. You should never delete the "visited" relations, because once a room is visited, it will remain that way.
\end{displayquote}
For Cooking World, the prompt is mostly the same for the first stage (looking for ingredients), with an additional LLM instance to identify closed containers, and their contents once opened. As described above, all found ingredients are mechanically picked up (hard-coded). 

For Action-gen, the prompt is simply a description of the simulation, providing as much information as specified in the above domain files. For Coin Collector, it is:
\begin{displayquote}
\small
You will play a game where your goal is to collect a coin. You need to move through rooms explore them. Sometimes, two rooms are connected by closed door that you need to open before you can go from one to another. You should also keep track of which room you have visited, and the direction at which you enter a room.

I will provide you with a description of the environment, and you will take one of the valid actions. Ready?
\end{displayquote}
For Cooking World, it is:
\begin{displayquote}
\small
You will play a game where your goal is to read a recipe, find ingredients, cook a meal, and eat the meal. The recipe includes the ingredients that you'll need to collect. The ingredients are scattered around rooms and may be found in containers. After you find the ingredients, you need to process them as required in the recipe. Here are how the ingredients are processed:

- slice: use a knife to slice the ingredient

- chop: use a knife to chop the ingredient

- dice: use a knife to dice the ingredient

- grill: use a toaster or a barbeque to cook the ingredient will grill it

- roast: use an oven to cook the ingredient will roast it

- fry: use a stove to cook the ingredient will fry it

You have to process the ingredients as specified in the recipe, otherwise you will fail. Once the ingredients are processed, you can cook the meal and eat the meal in the kitchen, so make sure you go back to the kitchen at that point.
             
Now, I will provide you with a description of the environment, and you will take one of the valid actions. Ready?
\end{displayquote}

\begin{figure}
    \centering
    \begin{lstlisting}
(define (domain environment)
  (:requirements :strips :typing :negative-preconditions :disjunctive-preconditions)
  (:types
    location
    direction
  )
  (:predicates
    (at ?loc - location)
    (visited ?loc - location)
    (connected ?loc1 - location ?loc2 - location ?dir - direction)
    (closed_door ?loc1 - location ?loc2 - location)
  )

  (:action move
    :parameters (?loc1 - location ?loc2 - location ?dir - direction)
    :precondition (and (at ?loc1) (connected ?loc1 ?loc2 ?dir) (not (closed_door ?loc1 ?loc2)))
    :effect (and (not (at ?loc1)) (at ?loc2))
  )

  (:action open_door
    :parameters (?loc1 - location ?loc2 - location)
    :precondition (and (at ?loc1) (closed_door ?loc1 ?loc2))
    :effect (not (closed_door ?loc1 ?loc2))
  )
)
    \end{lstlisting}
    \caption{Annotated domain file for Coin Collector.}
    \label{fig:coin_df}
\end{figure}

\begin{figure}
    \centering
    \begin{lstlisting}
(define (domain environment)
  (:requirements :strips :typing :negative-preconditions :disjunctive-preconditions)
  
  (:types
    ingredient container knife toaster stove oven barbeque - object
    location 
    direction
  )
  
  (:predicates
    (at ?loc - location)
    (obj_at ?obj - object ?loc - location)
    (visited ?loc - location)
    (connected ?loc1 - location ?loc2 - location ?dir - direction)
    (closed_door ?loc1 - location ?loc2 - location)
    
    (grilled ?ing - ingredient)
    (roasted ?ing - ingredient)
    (fried ?ing - ingredient)
    (chopped ?ing - ingredient)
    (sliced ?ing - ingredient)
    (diced ?ing - ingredient)
    (have ?obj - object)
  )

  (:action move
    :parameters (?loc1 - location ?loc2 - location ?dir - direction)
    :precondition (and (at ?loc1) (connected ?loc1 ?loc2 ?dir) (not (closed_door ?loc1 ?loc2)))
    :effect (and (not (at ?loc1)) (at ?loc2))
  )

  (:action open_door
    :parameters (?loc1 - location ?loc2 - location)
    :precondition (and (at ?loc1) (closed_door ?loc1 ?loc2))
    :effect (not (closed_door ?loc1 ?loc2))
  )
  
  (:action use_stove
    :parameters (?ing - ingredient ?loc - location ?sto - stove)
    :precondition (and (at ?loc) (obj_at ?sto ?loc) (have ?ing))
    :effect (fried ?ing)
  )
  
  (:action use_toaster
    :parameters (?ing - ingredient ?loc - location ?toa - toaster)
    :precondition (and (at ?loc) (obj_at ?toa ?loc) (have ?ing))
    :effect (grilled ?ing)
  )
  
  (:action use_oven
    :parameters (?ing - ingredient ?loc - location ?ove - oven)
    :precondition (and (at ?loc) (obj_at ?ove ?loc) (have ?ing))
    :effect (roasted ?ing)
  )
  
  (:action use_barbeque
    :parameters (?ing - ingredient ?loc - location ?bbq - barbeque)
    :precondition (and (at ?loc) (obj_at ?bbq ?loc) (have ?ing))
    :effect (grilled ?ing)
  )
  
  (:action chop
    :parameters (?ing - ingredient ?kni - knife)
    :precondition (and (have ?ing) (have ?kni))
    :effect (chopped ?ing)
  )
  
  (:action slice
    :parameters (?ing - ingredient ?kni - knife)
    :precondition (and (have ?ing) (have ?kni))
    :effect (sliced ?ing)
  )
  
  (:action dice
    :parameters (?ing - ingredient ?kni - knife)
    :precondition (and (have ?ing) (have ?kni))
    :effect (diced ?ing)
  )
)
    \end{lstlisting}
    \caption{Annotated domain file for Cooking World.}
    \label{fig:cooking_df}
\end{figure}

\section{Hyperparameters}
\label{app:hyperparameters}
For both simulations, we use the implementation from \citet{jansen2022textworldexpress}. For Coin Collector, we use the most complex setting supported by the system of 11 rooms with random connectivity, allowing up to 50 exploration steps. For Cooking World, we consider an easy setting with 2 rooms and 2 ingredients up to 20 steps and a hard setting of 5 rooms and 5 ingredients up to 50 steps. For both datasets, we vary the random random seed to generate randomize environment configurations, and use 0-9 as the development set, and 10-59 as the test set. 

For the choice of LLM, we consider \texttt{gpt-3.5-turbo-1106} (GPT 3.5 Turbo) and \texttt{gpt-4-1106-preview} (GPT 4 Turbo) across baseline methods (i.e., Action-gen, PDDL-gen, and PDDL-edit). We set the temperature to 1 to study stability. For Action-gen, we prompt the LLM with a full description of the simulation with the aim that a human player can learn to succeed. For the PDDL approaches, whenever the generated or edited problem cannot be solved by the domain file, indicating an error, the model is allowed to retry up to 5 times before failing the task. Models are provided with a hand-annotated domain file for each task contains permitted actions (e.g., move, open door, pick up) with well-defined types and predicates, parameters, pre-conditions, and effects. For the PDDL-edit setting, we prompt the LLM to generate edits (add, replace, and delete lines in the problem file) in JSON with the force-JSON flag set. The prompt of each method include a 1-shot demonstration of the output format (i.e., an action, a problem file, or edits to a problem file). See details of prompt design and domain file annotation in the Appendix~\ref{app:df}.

\section{Example Trajectories}
\label{app:trajectories}
A sample trajectory by GPT 4 Turbo with PDDL-edit on Cooking World (hard) is shown in Figure~\ref{fig:cooking_trajectory}. The navigation portion of the interaction (look around, open door, move) is inherited from Coin Collector.

\begin{figure*}
    \centering
    \begin{lstlisting}
< examine cookbook
> Gather all following ingredients and follow the directions to prepare this tasty meal.
Ingredients:
  black pepper, block of cheese, salt, red apple, yellow potato
Directions:
  slice the block of cheese, dice the red apple, chop the yellow potato, grill the yellow potato, prepare meal
< look around
> You are in the kitchen. In one part of the room you see a stove. There is also an oven. You also see a fridge that is closed. In another part of the room you see a counter that has a knife, a red apple, a raw yellow potato, and a cookbook on it. In one part of the room you see a kitchen cupboard that is closed. There is also a cutlery drawer that is closed. You also see a trash can that is closed. In another part of the room you see a dishwasher that is closed. In one part of the room you see a dining chair, that has nothing on it. 
To the South you see a closed plain door. To the West you see the corridor. 
< open fridge
> You open the fridge. The fridge contains a block of cheese.
< take block of cheese
> You take the block of cheese.
< take knife
> You take the knife.
< take red apple
> You take the red apple.
< take yellow potato
> You take the yellow potato.
< open kitchen cupboard
> You open the kitchen cupboard. It's empty inside.
< open cutlery drawer
> You open the cutlery drawer. It's empty inside.
< open trash can
> You open the trash can. It's empty inside.
< open dishwasher
> You open the dishwasher. It's empty inside.
< move west
> You are in the corridor. In one part of the room you see a key holder, that has nothing on it. There is also a shoe cabinet that is closed. You also see a umbrella stand, that has nothing on it. In another part of the room you see a hat rack, that has nothing on it. In one part of the room you see a coat hanger, that has nothing on it. 
To the North you see a closed wood door. To the East you see the kitchen. To the West you see a closed sliding patio door. 
< open shoe cabinet
> You open the shoe cabinet. It's empty inside.
< open door to north
> You open the wood door, revealing the bedroom. 
< move north
> You are in the bedroom. In one part of the room you see a dressing table, that has nothing on it. There is also a desk chair, that has nothing on it. You also see a desk, that has nothing on it. In another part of the room you see a chest of drawers that is closed. In one part of the room you see a wardrobe that is closed. There is also a night stand, that has nothing on it. You also see a bed, that has nothing on it. 
Through an open wood door, to the South you see the corridor. 
< open chest of drawers
> You open the chest of drawers. It's empty inside.
< open wardrobe
> You open the wardrobe. It's empty inside.
< move south
> You are in the corridor. (description omitted)
< open door to west
> You open the sliding patio door, revealing the backyard. 
< move west
> You are in the backyard. In one part of the room you see a barbeque. There is also a workbench, that has nothing on it. You also see a patio chair, that has nothing on it. In another part of the room you see a patio table, that has nothing on it. In one part of the room you see a clothes line, that has nothing on it. There is also a garden, that has nothing on it. 
Through an open sliding patio door, to the East you see the corridor. 
< move east
> You are in the corridor. (description omitted)
< move east
> You are in the kitchen. (description omitted)
< open door to south
> You open the plain door, revealing the pantry. 
< move south
> You are in the pantry. In one part of the room you see a folding chair, that has nothing on it. There is also a shelf that has some black pepper, and some salt on it. 
Through an open plain door, to the North you see the kitchen. 
< take black pepper
> You take the black pepper.
< take salt
> You take the salt.
< dice red apple
> You dice the red apple.
< slice block of cheese
> You slice the block of cheese.
< chop yellow potato
> You chop the yellow potato.
< move north
> You are in the kitchen. (description omitted)
Through an open plain door, to the South you see the pantry. To the West you see the corridor. 
< move west
> You are in the corridor. (description omitted)
< move west
> You are in the backyard. (description omitted)
< cook yellow potato in barbeque
> You grill the yellow potato with the barbeque.
< move east
> You are in the corridor. (description omitted)
< move east
> You are in the kitchen. (description omitted)
< prepare meal
> Adding the meal to your inventory.
< eat meal
> You eat the meal.  It is delicious.
    \end{lstlisting}
    \caption{An example trajectory performed by GPT 4 Turbo and PDDL-edit on Cooking World (hard).}
    \label{fig:cooking_trajectory}
\end{figure*}

\section{Generated Problem Files}
\label{app:example_pf}

\begin{figure}
    \centering
    \includegraphics[width=\columnwidth]{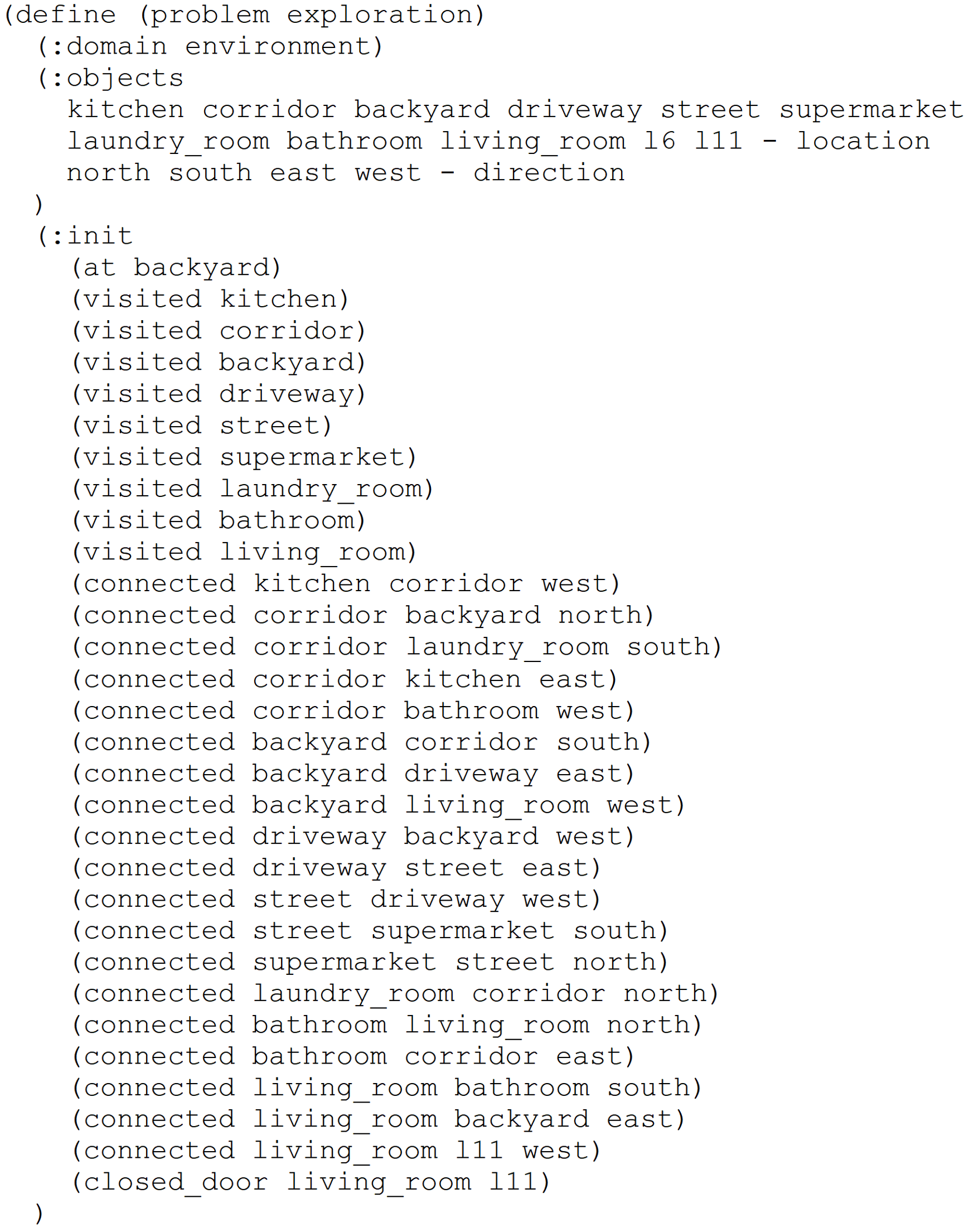}
    \caption{An example PDDL problem file learned throughout exploration in Coin Collector.}
    \label{fig:coin_pf}
\end{figure}

Figure~\ref{fig:coin_pf} demonstrates a PDDL problem file learned throughout exploration in Coin Collector that indicates the existence and connectivity of all rooms the agent has access to before finding the coin. 

\section{Results on the Development Set}
Table~\ref{tab:dev_results} is the counterpart of Table~\ref{tab:results} showcasing the results on the development set.

\begin{table*}[]
\centering
\small
\begin{tabular}{l|l|lll|lll}
\toprule
         & \multicolumn{1}{l|}{random} & \multicolumn{3}{c|}{GPT 3.5 Turbo} & \multicolumn{3}{c}{GPT 4 Turbo} \\ \midrule
         & & Action-gen  & PDDL-gen$^\dag$  & PDDL-edit$^\dag$ & Action-gen    & PDDL-gen$^\dag$   & PDDL-edit$^\dag$   \\
Coin    & 20\%  & 80\% & 30\% & 70\% & \textbf{90\%} & 50\% & 80\%    \\
Cooking (easy) & 0\% & 0\% &  100\% & 70\% & 10\% & 90\% & \textbf{100\%} \\
Cooking (hard) & 0\% & 0\% & 0\%  & 0\% & 0\% & 0\% & \textbf{50\%}            \\\bottomrule
\end{tabular}
\caption{The percentage where the agent succeeds by taking no more than the maximum steps on the development set. The $^\dag$ sign specifies methods under our proposed \ours methodology. }
\label{tab:dev_results}
\end{table*}

\end{document}